\documentclass[11pt]{article}
\usepackage[preprint]{acl}
\usepackage{times}
\usepackage{latexsym}
\usepackage[T1]{fontenc}
\usepackage[utf8]{inputenc}
\usepackage{microtype}
\usepackage{inconsolata}

\usepackage{graphicx}
\usepackage{amsmath}
\usepackage{multicol,multirow}
\usepackage{enumitem}
\usepackage{booktabs}
\usepackage{pifont}
\usepackage[table]{xcolor}

\title{From Human Intention to Action Prediction: Intention-Driven \\ End-to-End Autonomous Driving}

\author{
Huan Zheng\textsuperscript{1}\footnotemark[1], 
Yucheng Zhou\textsuperscript{1}\footnotemark[1], 
Tianyi Yan\textsuperscript{1}\footnotemark[1], 
Jiayi Su\textsuperscript{1}, 
Hongjun Chen\textsuperscript{1}, 
Dubing Chen\textsuperscript{1}, \\ 
\bf Xingtai Gui$^1$,
\bf Wencheng Han$^1$, 
\bf Runzhou Tao\textsuperscript{2}, 
\bf Zhongying Qiu\textsuperscript{2}, 
\bf Jianfei Yang\textsuperscript{2}, 
\bf Jianbing Shen\textsuperscript{1}\footnotemark[2] \\
$^1$SKL-IOTSC, CIS, University of Macau \\
$^2$Zhejiang ZEEKR Automobile Research \& Development Co., Ltd. \\
}

\begin{document}
\maketitle
\renewcommand{\thefootnote}{\fnsymbol{footnote}}
\footnotetext[1]{Equal Contribution.} 
\footnotetext[2]{Corresponding Author.}
\begin{abstract}
While end-to-end autonomous driving has achieved remarkable progress in geometric control, current systems remain constrained by a command-following paradigm that relies on simple navigational instructions. 
Transitioning to genuinely intelligent agents requires the capability to interpret and fulfill high-level, abstract human intentions. 
However, this advancement is hindered by the lack of dedicated benchmarks and semantic-aware evaluation metrics. 
In this paper, we formally define the task of Intention-Driven End-to-End Autonomous Driving and present Intention-Drive, a comprehensive benchmark designed to bridge this gap. 
We construct a large-scale dataset featuring complex natural language intentions paired with high-fidelity sensor data. 
To overcome the limitations of conventional trajectory-based metrics, we introduce the Imagined Future Alignment (IFA), a novel evaluation protocol leveraging generative world models to assess the semantic fulfillment of human goals beyond mere geometric accuracy. 
Furthermore, we explore the solution space by proposing two distinct paradigms: an end-to-end vision-language planner and a hierarchical agent-based framework. 
The experiments reveal a critical dichotomy where existing models exhibit satisfactory driving stability but struggle significantly with intention fulfillment.
Notably, the proposed frameworks demonstrate superior alignment with human intentions.
\end{abstract}

\section{Introduction}

\begin{figure*}[t]
    \centering
    \includegraphics[width=\linewidth]{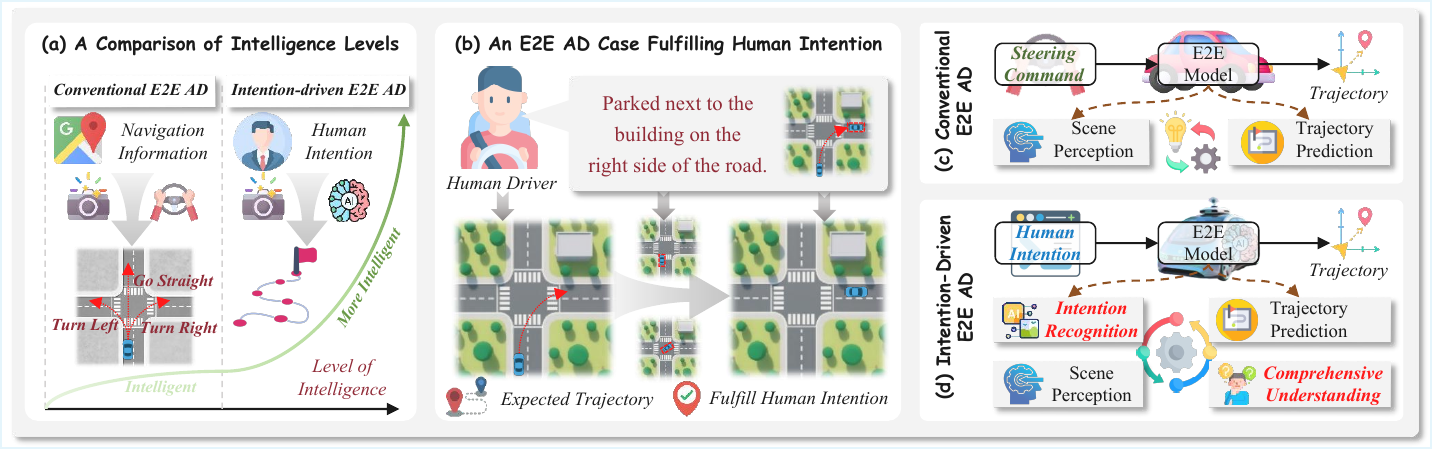}
    \vspace{-7mm}
    \caption{\small 
    \textbf{Conceptual illustration of the shift from conventional to intention-driven end-to-end autonomous driving.}
    (a) A comparison of intelligence levels, highlighting the leap from executing low-level navigational commands to understanding high-level, abstract human intentions.
    (b) An example scenario where the autonomous vehicle must interpret and fulfill the complex intention, which requires reasoning beyond simple steering command.
    (c) The architecture of a conventional end-to-end autonomous driving system, which acts as a command-follower by translating low-level steering commands and scene perception into a trajectory.
    (d) Our proposed paradigm for an intention-driven system, which functions as an intention-fulfiller.
    }
    \label{fig:motivation}
    \vspace{-4mm}
\end{figure*}

End-to-end autonomous driving (E2E AD) has recently gained significant traction, promising to overcome the limitations of traditional modular pipelines by learning complex driving policies directly from data~\cite{chen2024end, mao2023language, shao2024lmdrive, wu2025multi}. 
This approach has led to remarkable progress in vehicle control and navigation~\cite{hu2023planning, fu2025orion, ma2024learning}. 
However, as illustrated in Figure~\ref{fig:motivation}(a), these systems predominantly operate at a basic intelligence level. 
They function as mere command-followers that execute simple navigational instructions such as ``turn left'', ``turn right'', or ``go straight'', rather than as intelligent agents capable of interpreting human goals~\cite{hu2023planning}.
This fundamental limitation prevents current systems from achieving genuinely intelligent autonomy, which requires a paradigm shift from merely executing steering commands to understanding and fulfilling high-level, abstract human intentions. 
The leap from a command-follower to an intention-fulfiller represents a critical milestone in E2E AD intelligence, yet this transition is currently hindered by a significant gap.

The distinction between command-following and intention-fulfilling becomes apparent when examining complex real-world scenarios.
As shown in Figure \ref{fig:motivation}(b), a human driver can readily interpret and act upon an instruction like ``parked next to the building on the right side of the road''. 
This requires not just geometric path planning but also comprehensive scene understanding, spatial reasoning, and semantic interpretation. 
In contrast, as presented in Figure \ref{fig:motivation}(c), conventional E2E AD systems operate on a fundamentally different architectural principle, processing simple steering commands and scene perception to generate trajectories without the comprehension of the driver's underlying goal~\cite{hwang2024emma,pan2024vlp,liu2025hybrid}. 
The desired paradigm, as shown in Figure \ref{fig:motivation}(d), would function as an intention-fulfiller, directly taking high-level human intention as input and requiring a comprehensive understanding that fuses intention recognition with scene perception to generate appropriate driving actions.

Recent research attempts to incorporate Large Language Models (LLMs) into autonomous driving frameworks, leveraging their semantic understanding capabilities~\cite{han2025dme, sima2024drivelm, yuan2024rag, yang2023llm4drive, yang2025drivemoe, yang2025trajectory}. 
However, a critical issue has emerged: when these models undergo task-specific training using fixed, limited sets of query templates rather than diverse natural instructions, their ability to interpret human intentions degrades~\cite{li2025recogdrive, fu2025orion, zeng2025futuresightdrive}. 
Although built upon LLMs that possess strong language understanding capabilities, these models become overly specialized for trajectory prediction at the expense of semantic fidelity, ultimately failing to accurately interpret complex human instructions like those shown in Figure \ref{fig:motivation}(b). 
This phenomenon represents a significant obstacle in the development of intention-aware AD systems.

The field is further hampered by the absence of standardized evaluation frameworks that can properly measure a system's ability to understand and fulfill human intentions. 
Current evaluation metrics~\cite{xu2024drivegpt4, hu2023planning, Dauner2024NEURIPS} focus primarily on the geometric accuracy of trajectory prediction, which fails to capture whether the vehicle has actually satisfied the human's underlying goal. 
Without appropriate benchmarks to quantify this crucial capability, progress toward intention-driven autonomous driving remains unmeasurable and therefore difficult to achieve. 
This critical gap has persisted despite the increasing sophistication of E2E AD systems.

To address these fundamental challenges, we formulate the task of intention-driven end-to-end autonomous driving. 
We further introduce Intention-Drive, the first comprehensive benchmark specifically designed to evaluate an AD system's ability to translate high-level human intentions into safe and precise driving actions. 
Using this benchmark, we develop and evaluate two distinct methodological frameworks to tackle the intention-driven task.
Through extensive experiments, we analyze the capability of current state-of-the-art models, revealing a critical dichotomy between basic driving competency and intention understanding.

The key contributions of this work are fourfold:
\begin{itemize}[leftmargin=*, itemsep=1pt, topsep=1pt, partopsep=1pt, parsep=1pt]
    \item We formally define \emph{Intention-Driven End-to-End Autonomous Driving}, a new problem setting that requires models to ground high-level, abstract human intentions into driving trajectories.
    \item We introduce \emph{Intention-Drive}, the first comprehensive benchmark for this task, featuring a large-scale dataset with natural language intentions and a novel evaluation protocol based on \emph{Imagined Future Alignment (IFA)}, which assesses semantic goal fulfillment beyond geometric accuracy.
    \item We present two baseline frameworks: an end-to-end vision-language planner that directly maps inputs to trajectories, and a hierarchical agent-based framework that decomposes intention understanding and vehicle action.
    \item Our experiments uncover a fundamental disconnect between geometric driving competence and semantic intention in existing models, while our approach achieves substantially improved alignment with human intentions.
\end{itemize}

\section{Related Work}
End-to-end (E2E) autonomous driving replaces the traditional modular pipeline with a single, jointly optimized model that maps sensor inputs directly to driving commands or trajectories~\cite{chen2024end, zhang2025vldrive}. 
Early approaches established strong baselines by unifying perception, prediction, and planning~\cite{hu2023planning, jiang2023vad}, while subsequent works incorporated uncertainty modeling~\cite{chen2024vadv2} and generative paradigms such as diffusion models~\cite{zheng2024genad, liao2025diffusiondrive}. 
Despite encouraging open-loop results, E2E systems often struggle in interactive closed-loop settings, exhibiting brittleness and suboptimal behaviors~\cite{jia2024bench2drive, fu2025orion}. 
To address these limitations, recent studies increasingly explore Vision-Language-Action (VLA) models to enhance robustness and high-level decision-making~\cite{li2025recogdrive, zhou2025opendrivevla, chen2025drivinggpt}.
VLA models integrate multi-modal perception, language understanding, and action generation, enabling agents to interpret high-level instructions and execute grounded behaviors~\cite{ma2024survey, driess2023palm}. 
In autonomous driving, prior work emphasizes human-like cognitive structures and explicit reasoning for improved interpretability and control~\cite{wang2025cogad, lu2025real}, including Chain-of-Thought reasoning~\cite{yuan2025autodrive, li2025recogdrive} and policy optimization via RL or DPO~\cite{jiang2025alphadrive, fang2025corevla}. 
Recent advances further incorporate active perception to reduce uncertainty during decision-making~\cite{zheng2025driveagent}. 
A full version is provided in Appendix~\ref{app:related_work}.

\section{Intention-Drive Benchmark}
\begin{figure*}[t]
    \centering
    \includegraphics[width=\linewidth]{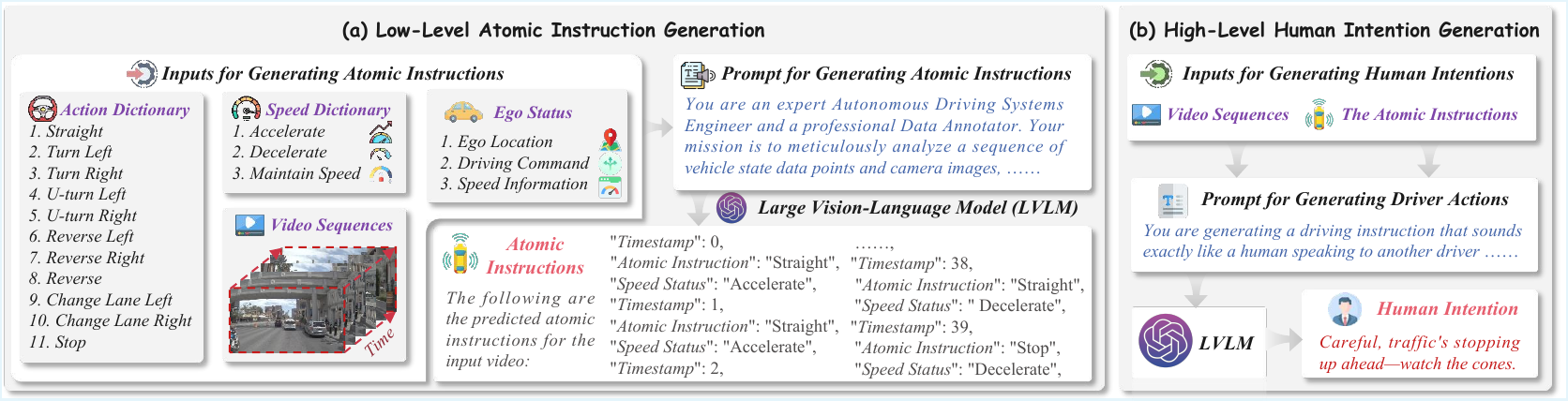}
    \vspace{-7mm}
    \caption{\small \textbf{The pipeline of hierarchical dataset construction.} We utilize Large Vision-Language Models (LVLMs) to annotate scenarios across three abstraction levels. (a) Low-Level Atomic Instruction Generation infers fine-grained motion primitives from vehicle states and video sequences. (b) High-Level Human Intention Generation produces natural, context-aware human intentions by reasoning over the visual scene and vehicle dynamics.}
    \label{fig:dataset}
    \vspace{-4mm}
\end{figure*}

\subsection{Task Definition}
We formally define the task of intention-driven end-to-end autonomous driving. 
The primary objective is to generate a safe, feasible, and contextually appropriate driving trajectory that semantically satisfies a human intention. 
This task requires an agent to move beyond simple navigational command following and instead perform complex reasoning that grounds linguistic concepts.

Formally, let the state of the environment at timestep $t$ be represented by a set of sensor observations $O_t$, consisting solely of multi-view camera images $\{I_t^1, \dots, I_t^N\}$.
Given a history of observations over a time horizon $H$, denoted as $\mathcal{O}_{t-H:t}$, and a high-level human intention articulated as a natural language instruction $I_{\text{lang}}$, the goal is to learn a policy $\pi$ that predicts a future driving trajectory $T_{t+1:t+K}$. 
The trajectory consists of $K$ waypoints in the bird's-eye-view (BEV) coordinate frame, $T_{t+1:t+K} = \{p_{t+1}, \dots, p_{t+K}\}$, where each waypoint is defined as $p_k = (x_k, y_k)$. The policy is parameterized by $\theta$ and can be expressed as:
\begin{equation}
    T_{t+1:t+K} = \pi(\mathcal{O}_{t-H:t}, I_{\text{lang}} | \theta).
\end{equation}
Specifically, the input is defined as a tuple $(\mathcal{O}_{t-H:t}, I_{\text{lang}})$, where $\mathcal{O}_{t-H:t}$ represents a sequence of visual data capturing the history of the scene, and $I_{\text{lang}}$ denotes a free-form natural language string describing human intention. Based on these inputs, the model predicts the output future trajectory $T_{t+1:t+K}$ for the ego-vehicle, which is formalized as a sequence of $K$ 2D waypoints.

This task presents several fundamental challenges that distinguish it from conventional end-to-end driving. 
\ding{182} It demands robust semantic grounding, requiring the model to associate abstract linguistic concepts with their corresponding physical entities and spatial relationships in the 3D world. 
\ding{183} It necessitates sophisticated compositional reasoning to deconstruct complex instructions into a sequence of executable driving maneuvers. 
\ding{184} Evaluating success cannot rely solely on geometric metrics to a ground-truth trajectory, as multiple distinct paths could validly fulfill the same intention.

\subsection{Data Construction}
Existing datasets \cite{Dauner2024NEURIPS, jia2024bench2drive, nuscenes2019} primarily focus on trajectory prediction from navigational commands and sensor data, lacking the rich and abstract language that characterizes human intentions. 
To bridge this gap, we introduce Intention-Drive, a benchmark specifically designed for intention-driven end-to-end autonomous driving. 

The creation of the Intention-Drive dataset follows a multi-stage pipeline designed to generate realistic and diverse intention-action pairs. 
Our process leverages the OpenScene dataset ~\cite{openscene2023} and employs advanced Large Language Models (LLMs) for language annotation. 
The pipeline consists of the following key steps:

\noindent\textbf{Foundational Scenario Curation.}
The construction of our benchmark commences with the rigorous curation of driving scenarios sourced from the large-scale OpenScene dataset~\cite{openscene2023, sima2023_occnet}.
From this corpus, we strategically select and filter a subset of scenarios, comprising 18,765 scenes for training and 1,959 for evaluation.
For each scenario, we extract a temporal sequence of sensor observations paired with the ego-vehicle's future trajectory.

\noindent\textbf{Hierarchical Dataset Construction.}
We devised a hierarchical annotation strategy powered by an advanced Large Vision-Language Model (LVLM). 
Specifically, we employ GPT-5.2 as the core annotation engine, leveraging its state-of-the-art capabilities in multimodal reasoning and semantic alignment to parse complex driving scenarios.
This process synthesizes two distinct levels of linguistic instructions for each driving scenario, as illustrated in Figure~\ref{fig:dataset}. 
The detailed pipeline for obtaining these instructions is described as follows:
\begin{itemize}[leftmargin=*, itemsep=1pt, topsep=1pt, partopsep=1pt, parsep=1pt]
    \item \textit{Low-Level Atomic Instructions}: We first define the most granular, standardized driving primitives representing the fundamental actions of the ego vehicle. 
    For each timestamp, a VLM is employed to infer the primary maneuver from a predefined vocabulary of atomic actions, such as \texttt{Straight}, \texttt{Turn Left}, \texttt{Change Lane Right}, and \texttt{Stop}. 
    Concurrently, a speed status is assigned, reflecting the vehicle's instantaneous speed profile. 
    This initial layer ensures a precise and unambiguous representation of the vehicle's movement at the most basic level, forming the foundational understanding for subsequent high-level generations.
    \item \textit{High-Level Human Intentions}: Next, leveraging the comprehensive understanding derived from low-level atomic instructions, alongside the sensor inputs, these instructions capture natural language utterances from a passenger's perspective. 
    These represent abstract goals or observations that require a higher degree of contextual reasoning and semantic interpretation. 
    This high level focuses on mirroring the intuitive way humans communicate driving intentions, often blending direct commands with observations or warnings.
\end{itemize}

\noindent\textbf{Data Verification and Refinement.}
To ensure annotation quality, we implement a rigorous verification pipeline where the VLM evaluates the logical coherence and semantic alignment between instruction levels. 
Any identified discrepancies trigger a human-in-the-loop refinement process for manual correction to guarantee the reliability.

\subsection{Evaluation Protocol}
To rigorously assess autonomous agents, we propose a comprehensive evaluation framework that decouples basic driving competency from high-level intention fulfillment. This protocol integrates a standardized metric for driving quality (following \cite{hu2023planning,nuscenes2019}) with a novel generative approach for semantic verification.

\subsubsection{Geometric Metrics} Before an agent can fulfill complex intentions, it must ensure basic driving safety and geometric stability. We employ two standard metrics to evaluate this fundamental capability:

\noindent\textit{Average Displacement Error (ADE)}: This metric measures the geometric fidelity of the predicted trajectory. It is calculated as the average $L_2$ distance between the predicted waypoints and the ground truth trajectory over the time horizon. While a low ADE indicates stable motion generation akin to the human demonstration, it is insufficient alone for intention evaluation, as multiple distinct paths may validly fulfill the same abstract intention.

\noindent\textit{Collision Rate (CR)}: This serves as a hard safety constraint. It is defined as the percentage of test scenarios where the ego-vehicle collides with any obstacles or road boundaries. A collision represents a critical failure of the driving system.

\subsubsection{Imagined Future Alignment}
While the geometric metrics ensure kinematic safety, it remains blind to the semantic nuances of human intention. 
To bridge this gap, we introduce Imagined Future Alignment (IFA), a novel evaluation paradigm that leverages a generative world model to explicitly reason about the future consequences of planned actions.

\noindent\textbf{Conditional Future Hallucination.}
Instead of evaluating abstract trajectory coordinates, we map the agent's plan back into the visual domain. 
We employ a pre-trained generative world model~\cite{gao2024vista}, denoted as $\mathcal{W}$, which serves as a neural simulator. 
Given the current visual observation sequence $\mathcal{O}_t$ and the agent's predicted trajectory $\mathcal{T}_{\text{pred}}$, the world model hallucinates a photo-realistic future video clip $\hat{\mathcal{V}}_{\text{future}}$:
\begin{equation}
    \hat{\mathcal{V}}_{\text{future}} = \mathcal{W}(\mathcal{O}_t, \mathcal{T}_{\text{pred}}).
\end{equation}
This process effectively translates the numerical driving plan into a perceptible visual narrative, enabling a more grounded semantic evaluation.

\noindent\textbf{Dual-Aspect Semantic Reasoning.}
We utilize a VLM as a \textit{Semantic Judge} to verify the alignment between the hallucinated future $\hat{\mathcal{V}}_{\text{future}}$ and the high-level human intention $\mathcal{I}_{\text{lang}}$. To capture both the execution quality and the final outcome, we decompose the evaluation into two distinct scores:

\noindent\textit{Process Fidelity Score} ($\mathcal{S}_{\text{proc}} \in [0, 1]$): This continuous metric measures the semantic consistency of the driving behavior throughout the entire video duration. It reflects how well the agent's driving style, speed, and interaction patterns match the descriptive aspects of the intention.

\noindent\textit{Goal Completion Score} ($\mathcal{S}_{\text{goal}} \in \{0, 1\}$): This binary metric assesses the success of the final state. The Judge examines the end of the hallucinated sequence to determine if the core objective (e.g., reaching a specific destination, parking in the correct slot) has been definitively achieved.

\noindent\textbf{IFA Calculation.}
The final IFA score for a given scenario is computed as the product of the goal completion and the process fidelity. This design ensures that a scenario is only considered valid if the goal is achieved, while the score magnitude rewards precise adherence to behavioral instructions:
\begin{equation}
    \text{IFA} = \mathcal{S}_{\text{goal}} \times \mathcal{S}_{\text{proc}}
\end{equation}
By integrating the world model, IFA transcends traditional geometric metrics, providing a verifiable and interpretable measure of how well an autonomous agent understands and reasons about the future in the context of human commands.

\begin{figure*}[t]
    \centering
    \includegraphics[width=\linewidth]{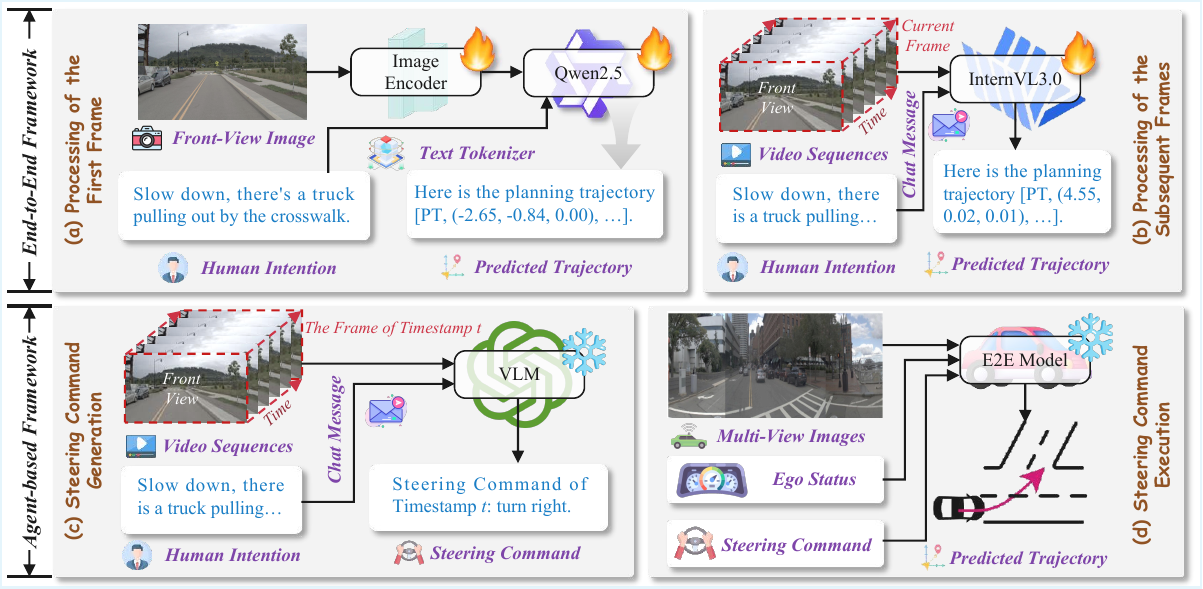}
    \vspace{-6mm}
    \caption{\small 
    \textbf{Overview of the proposed methodological frameworks.}
    Top: The End-to-End Framework ((a)-(b)) employs InternVL3.0 to fuse visual encodings with high-level natural language intentions, directly regressing driving trajectories via a unified LLM backbone.
    Bottom: The Agent-Based Framework ((c)-(d)) adopts a hierarchical structure to bridge the semantic gap. A VLM first acts as a reasoning agent to translate abstract intentions into discrete steering commands, which subsequently condition a conventional E2E AD model for execution.
  }
    \label{fig:solution}
    \vspace{-4mm}
\end{figure*}

\section{Method}
To systematically address the challenges of interpreting and executing high-level human intentions, we explore the solution space through two distinct methodological paradigms. 
We first introduce a unified end-to-end framework that leverages a Large Vision-Language Model to directly regress driving trajectories from multi-modal visual-linguistic inputs. 
Subsequently, we present a hierarchical agent-based framework that decouples the task into semantic command generation and kinematic execution, explicitly bridging the gap between abstract intention and control.

\subsection{End-to-End Framework}
To systematically evaluate the proposed Intention-Drive benchmark, we propose a strong end-to-end baseline. 
Unlike traditional E2E models that rely on simple steering commands, the intention-fulfilling task demands a model capable of deep semantic grounding, mapping abstract linguistic concepts to specific visual features and geometric actions. 
To this end, we adopt InternVL3.0-2B~\cite{zhu2025internvl3} as our foundational architecture. 
The overall pipeline is illustrated in Figure~\ref{fig:solution}(a)-(b).

\noindent\textbf{Visual-Linguistic Encoding.}
Perceiving fine-grained scene details is a prerequisite for understanding complex human intentions. 
The proposed E2E framework utilizes a Vision Transformer (ViT) as the visual encoder $\Phi_{\text{vis}}$ to process the sequence of visual observations $\mathcal{O}_{1:t}$. 
To preserve spatial fidelity while adapting to the LLM's input space, the extracted features undergo a Pixel Unshuffle operation followed by an MLP projector. 
Formally, the visual tokens $\mathbf{Z}_{\text{vis}}$ are obtained as:
\begin{equation}
    \mathbf{Z}_{\text{vis}} = \text{MLP}(\text{PixelUnshuffle}(\Phi_{\text{vis}}(\mathcal{O}_{t-H:t}))).
\end{equation}
Simultaneously, the high-level human intention $I_{\text{lang}}$ is processed by a text tokenizer to yield the linguistic tokens:
\begin{equation}
    \mathbf{Z}_{\text{lang}} = \text{Tokenizer}(I_{\text{lang}}). 
\end{equation}
These visual and textual tokens are concatenated to form a unified multimodal input sequence $\mathbf{Z}_{\text{in}} = [\mathbf{Z}_{\text{vis}}, \mathbf{Z}_{\text{lang}}]$, projecting the driving scene and user intention into a shared embedding space.

\noindent\textbf{LLM-Based Reasoning.}
The core decision-making module is powered by Qwen2.5~\cite{bai2025qwen2}, a SoTA LLM parameterized by $\Theta_{\text{E2E}}$. By initializing with weights from ReCogDrive~\cite{li2025recogdrive}, the model inherits a preliminary understanding of driving physics. 

\noindent\textbf{Trajectory Generation.}
We formulate the planning problem as an auto-regressive token generation task. 
Instead of regressing numerical values directly, the model predicts the future trajectory $T_{t+1:t+K}$ in a textual format. The trajectory is serialized into a sequence of coordinate tokens $S = (s_1, s_2, \dots, s_M)$, structured as ``Here is the planning trajectory [PT, $(x_1, y_1), \dots, (x_K, y_K)$]''.

Mathematically, the model maximizes the likelihood of the target token sequence $S$ given the multimodal inputs. The joint probability is factorized into a product of conditional probabilities:
\begin{equation}
    \begin{split}
        P(S | \mathcal{O}_{t-H:t}, I_{\text{lang}}; \Theta_{\text{E2E}}) & \\
        = \prod_{i=1}^{M} P(s_i | s_{<i}, \mathbf{Z}_{\text{vis}}&, \mathbf{Z}_{\text{lang}}; \Theta_{\text{E2E}}),
    \end{split}
\end{equation}
where $s_{<i}$ denotes the history of generated tokens. During inference, the generated text sequence $S$ is parsed to extract the waypoints $\{p_{t+1}, \dots, p_{t+K}\}$ in the BEV coordinate system, constituting the final executable trajectory $T_{t+1:t+K}$.

\subsection{Agent-Based Framework}
Distinct from the architecture described above, the agent-based framework seeks to repurpose established conventional E2E models for the intention-driven paradigm. 
Since these models act primarily as command-followers conditioned on simple navigation signals, we introduce a hierarchical decomposition to bridge the semantic gap between abstract language and vehicle control. 
As illustrated in Figure~\ref{fig:solution}(c)-(d), this framework factorizes the complex reasoning process into two sequential phases: Command Generation via a VLM agent, and Trajectory Prediction via a standard planner.

\noindent\textbf{Command Generation.}
The objective of this phase is to translate the abstract human intention $I_{\text{lang}}$ into a standardized steering command that acts as a deterministic interface for the downstream planner. 
We define a discrete action space $\mathcal{C}$, including ``turn left'', ``turn right'' and ``forwrad''.

We employ a Vision-Language Model (VLM) as the reasoning agent to bridge the gap. 
The VLM processes the sequence of visual observations $\mathcal{O}_{t-H:t}$ and the natural language intention $I_{\text{lang}}$ to infer the optimal command $c_t$:
\begin{equation}
    c_t = \mathop{\arg\max}_{c \in \mathcal{C}} P(c | \mathcal{O}_{t-H:t}, I_{\text{lang}}; \theta_{\text{agent}}),
\end{equation}
where $\theta_{\text{agent}}$ represents the parameters of the reasoning agent. By explicitly predicting $c_t$, the system resolves the semantic ambiguity of the high-level intention before kinematic planning begins.

\noindent\textbf{Trajectory Prediction.}
In the second phase, we leverage an off-the-shelf conventional E2E driving model, denoted as $\pi_{\text{E2E}}$.
Unlike the end-to-end framework that directly regresses coordinates from language embeddings, this module operates in the traditional command-conditional mode.

Receiving the command $c_t$ generated by the agent, along with the sensory context $\mathcal{O}_{t-H:t}$, the model generates the future trajectory $T_{t+1:t+K}$:
\begin{equation}
    T_{t+1:t+K} = \pi_{\text{E2E}}(\mathcal{O}_{t-H:t}, c_t | \phi_{\text{drive}}),
\end{equation}
where $\phi_{\text{drive}}$ represents the pre-trained weights of the driving backbone. 

Mathematically, this agent-based approach factorizes the complex distribution of intention-driven trajectories into a chain of reasoning and execution:
\begin{equation}
    P(T | \mathcal{O}, I_{\text{lang}}) \approx \sum_{c \in \mathcal{C}} \underbrace{P(T | \mathcal{O}, c)}_{\text{Execution}} \cdot \underbrace{P(c | \mathcal{O}, I_{\text{lang}})}_{\text{Reasoning}}.
\end{equation}
This factorization allows us to combine the advanced semantic reasoning capabilities of VLMs with the trajectory generation of established E2E models, effectively transforming them from command-followers to intention-fulfillers.

\section{Experiments}

\subsection{Implementation Details}
We utilize InternVL3.0-2B~\cite{zhu2025internvl3}, initializing it with the pre-trained weights from the first stage of ReCogDrive~\cite{li2025recogdrive}. 
The model is fine-tuned using the AdamW optimizer with a learning rate of $2 \times 10^{-5}$ and a weight decay of $0.05$. We employ a cosine learning rate scheduler with a warmup ratio of $0.03$. 
The training is conducted for 1 epoch with a total batch size of 4 distributed across 4 GPUs. 
To process high-resolution driving scenes, we utilize a dynamic image size strategy with a base resolution of $448 \times 448$ and a maximum of 16 dynamic patches. 
For parameter efficiency, we apply Low-Rank Adaptation (LoRA) with a rank of $r=32$ to both the vision backbone and the LLM. 
The training process is optimized using DeepSpeed with BF16 precision and gradient checkpointing.

\begin{table}[t]\small
\centering
\begin{tabular}{l|cc|c}
\toprule
\multirow{2}{*}{\textbf{Method}} & \multicolumn{2}{c|}{\textbf{Geometric Metrics}} & \multicolumn{1}{c}{\textbf{Intention}} \\
& ADE (m)  & Coll. (\%)  & IFA (\%) \\
\midrule
\rowcolor{gray!15} \multicolumn{4}{l}{\textit{\textbf{General LLM}}} \\
GPT 5.2 & 1.78 & 0.53 & 23.3 \\
Gemini 3 Flash & 2.14 & 0.70 & 14.2 \\
Gemini 3 Pro & 1.58 & 0.42 & 26.8 \\
\midrule
\rowcolor{gray!15} \multicolumn{4}{l}{\textit{\textbf{Agent-based framework}}} \\
GPT5+RD & 1.46 & 0.41 & 23.3 \\
Gemini 3 Pro+RD & 1.46 & 0.34 & 24.5 \\
\midrule
\rowcolor{gray!15} \multicolumn{4}{l}{\textit{\textbf{End-to-end framework}}} \\
InternVL3.0-2B & \bf 0.70 & 0.26 & 33.7 \\
InternVL3.0-2B* & 0.72 & \bf 0.25 & \bf 35.6 \\
\bottomrule
\end{tabular}
\vspace{-2mm}
\caption{\small \textbf{Main results on Intention-Drive.} * denotes fine-tuning initialized with weights of ReCogDrive.}
\label{tab:main_results}
\vspace{-5mm}
\end{table}

\begin{table}[t]\small
\centering
  \setlength{\tabcolsep}{4pt}
\begin{tabular}{l|ccc|ccc}
\toprule
\multirow{2}{*}{\textbf{Method}} & \multicolumn{3}{c|}{\textbf{ADE (m)}} & \multicolumn{3}{c}{\textbf{Coll. (\%)}} \\
& 1s & 2s  & 3s & 1s & 2s  & 3s  \\
\midrule
\rowcolor{gray!15} \multicolumn{7}{l}{\textit{\textbf{General LLM}}} \\
GPT 5.2 & 0.97 & 1.75 & 2.61 & 0.19 & 0.43 & 0.99  \\
Gemini 3 Flash & 1.10 & 2.08 & 3.24 & 0.23 & 0.66 & 1.22  \\
Gemini 3 Pro & 0.51 & 1.41 & 2.83 & 0.19 & 0.35 & 0.71\\
\midrule
\rowcolor{gray!15} \multicolumn{7}{l}{\textit{\textbf{Agent-based framework}}} \\
GPT5+RD & \bf 0.24 & 1.22 & 2.92 & 0.13 & 0.32 & 0.77 \\
Gemini 3 Pro+RD & \bf 0.24 & 1.22 & 2.92 & \bf 0.09 & 0.29 & 0.65  \\
\midrule
\rowcolor{gray!15} \multicolumn{7}{l}{\textit{\textbf{End-to-end framework}}} \\
InternVL3.0-2B & 0.29 & \bf 0.65 & \bf 1.16 & 0.15 & \bf 0.17 & 0.46 \\
InternVL3.0-2B* & 0.26 & 0.70 & 1.17 & 0.14 & \bf 0.17 & \bf 0.45 \\
\bottomrule
\end{tabular}
\vspace{-2mm}
\caption{\small \textbf{Performance analysis across diverse time ranges.} * denotes fine-tuning initialized with weights of ReCogDrive.}
\label{tab:time_range}
\vspace{-5mm}
\end{table}

\subsection{Main Results}
We evaluate all models on the Intention-Drive benchmark. 
Table~\ref{tab:main_results} presents the quantitative results, revealing a critical dichotomy between driving competency and intention understanding.

\begin{figure*}[t]
    \centering
    \includegraphics[width=\linewidth]{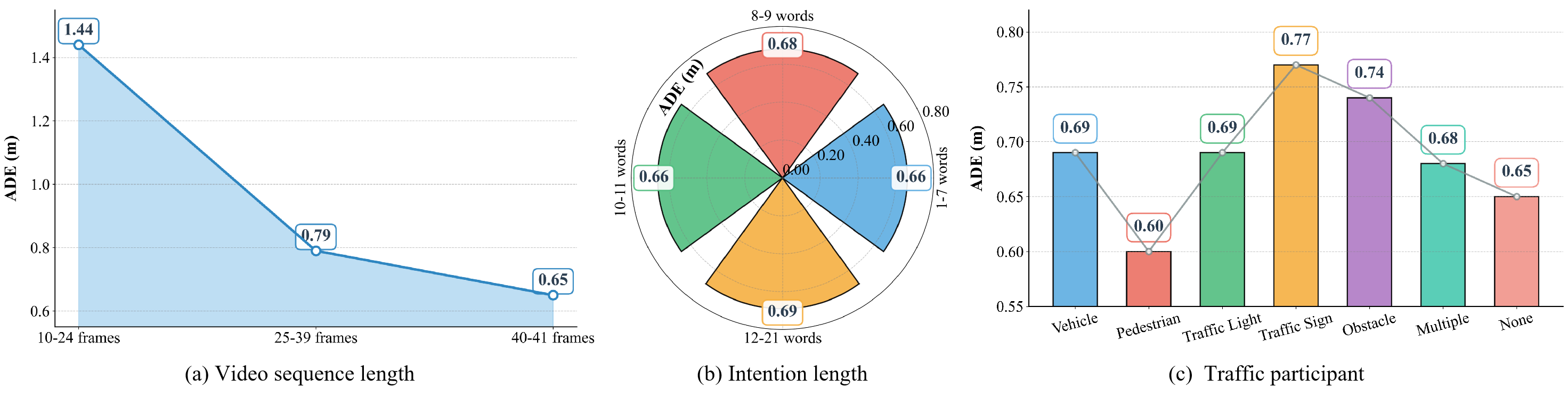}
    \vspace{-7mm}
    \caption{\small \textbf{Performance analysis for InternVL3.0-2B*.} (a) Analysis of planning performance across varying video sequence lengths. (b) Performance analysis of Intention lengths. (c) Performance analysis across different traffic participants.
    }
    \vspace{-4mm}
    \label{fig:performance_analysis}
\end{figure*}

\noindent\textbf{Safety and Geometric Stability.} 
As evidenced by the geometric metrics in Table~\ref{tab:main_results}, General LLMs struggle to generate kinematically feasible trajectories solely from visual-linguistic inputs. Models such as Gemini 3 Flash and GPT 5.2 exhibit high ADE of 2.14m and 1.78m, respectively, along with elevated collision rates. This indicates that while LLMs possess strong semantic reasoning, they lack the spatial grounding required to adhere to the physical constraints of driving scenes, often hallucinating unsafe paths. 
The agent-based frameworks improve stability slightly by decoupling reasoning from execution, yet they still lag behind the integrated approach. 
In contrast, the end-to-end framework demonstrates superior driving stability. 
The InternVL3.0-2B model achieves a remarkable ADE of 0.70m and a collision rate of 0.26\%, reducing the trajectory error by approximately 55\% compared to the best-performing General LLM. 
This confirms that end-to-end training effectively aligns high-level linguistic concepts with precise, physically grounded control policies.

\noindent\textbf{Intention Fulfillment Gap.}
Crucially, geometric precision does not guarantee semantic alignment with human goals. 
The proposed IFA metric highlights this disparity.
Although agent-based frameworks leverage the reasoning power of LLMs, their performance plateaus at an IFA of roughly 24\%. 
This limitation stems from the information bottleneck inherent in the hierarchical design: compressing complex, abstract intentions into discrete steering commands results in a significant loss of semantic nuance.
Conversely, the end-to-end framework establishes a direct mapping from inputs to driving trajectories, preserving the rich semantic information embedded in the human intentions. Consequently, the InternVL3.0-2B model achieves an IFA of 33.7\%, which further improves to 35.6\% when initialized with weights from ReCogDrive. 
This result empirically validates our hypothesis that a unified, intention-driven paradigm is essential for bridging the gap between understanding a command and accurately fulfilling it in complex real-world scenarios.

\subsection{Analysis of Time Range}
We investigate the temporal stability of the generated trajectories by decomposing the planning performance across prediction horizons of 1, 2, and 3 seconds, as shown in Table \ref{tab:time_range}.
While error accumulation is inevitable over longer durations due to increasing environmental uncertainty, the fine-tuned InternVL3.0-2B exhibits significantly superior robustness, maintaining a low ADE of 1.16m at the 3-second horizon.
In contrast, other frameworks suffer from severe spatial drift, with greater errors.
This demonstrates that our end-to-end fine-tuning strategy effectively grounds abstract intentions into precise, temporally consistent trajectory.

\subsection{Analysis of Video Length}
We study the impact on planning performance by categorizing test scenarios based on video sequence length, as illustrated in Figure~\ref{fig:performance_analysis}(a).
The results indicate that the model achieves superior performance on longer video sequences, as evidenced by the decreasing ADE.
We attribute this improvement to the stabilization of planning over time: while the initial phase suffers from kinematic uncertainty, extended sequences provide richer temporal context, allowing the model to rectify early deviations and generate more consistent trajectories.

\subsection{Analysis of Intention Length}
To evaluate the model's robustness to linguistic complexity, we analyze the planning performance across different intention lengths, as illustrated in Figure~\ref{fig:performance_analysis}(b). 
The results demonstrate that the model maintains a stable performance with the ADE fluctuating narrowly between 0.66m and 0.69m, regardless of the token count. 
This indicates that our framework effectively captures the semantic essence of high-level intentions, handling both concise commands and elaborate descriptions without significant performance degradation.

\subsection{Analysis of Traffic Participants}
To evaluate the model's sensitivity to specific semantic concepts, we categorize the validation scenarios based on the traffic participants explicitly referenced within the provided human intention.
As illustrated in Figure \ref{fig:performance_analysis}(c), the model exhibits the lowest ADE when the intention involves dynamic agents such as ``Vehicle'' and ``Pedestrian'', benefiting from their visual saliency and predictable kinematics.
In contrast, a performance drop is observed for intentions citing static regulatory elements like ``Traffic Sign'' and ``Traffic Light'', suggesting that grounding abstract linguistic instructions to small-scale, stationary visual cues requires more fine-grained reasoning than large objects.

\section{Conclusion}
This paper formalizes the transition in end-to-end autonomous driving from a command-following paradigm to an intention-driven intelligence level capable of fulfilling abstract human goals. 
To support this new task, we introduce the Intention-Drive benchmark, comprising a large-scale dataset and a generative evaluation protocol termed imagined future alignment to assess semantic goal fulfillment. 
We explore the solution space by proposing two distinct methodological paradigms: an end-to-end vision-language planner for direct trajectory prediction and a hierarchical agent-based framework that decouples intention reasoning from command execution. 
The experiments reveal a significant performance gap in existing models between driving stability and intention fulfillment, while demonstrating that our proposed frameworks achieve superior alignment with complex human instructions.

\bibliography{custom}

@article{chen2024vadv2,
  title={Vadv2: End-to-end vectorized autonomous driving via probabilistic planning},
  author={Chen, Shaoyu and Jiang, Bo and Gao, Hao and Liao, Bencheng and Xu, Qing and Zhang, Qian and Huang, Chang and Liu, Wenyu and Wang, Xinggang},
  journal={arXiv preprint arXiv:2402.13243},
  year={2024}
}

@inproceedings{hu2023planning,
  title={Planning-oriented autonomous driving},
  author={Hu, Yihan and Yang, Jiazhi and Chen, Li and Li, Keyu and Sima, Chonghao and Zhu, Xizhou and Chai, Siqi and Du, Senyao and Lin, Tianwei and Wang, Wenhai and others},
  booktitle={Proceedings of the IEEE/CVF conference on computer vision and pattern recognition},
  year={2023}
}

@article{wu2025multi,
  title={Multi-agent autonomous driving systems with large language models: A survey of recent advances, resources, and future directions},
  author={Wu, Yaozu and Li, Dongyuan and Chen, Yankai and Jiang, Renhe and Zou, Henry Peng and Huang, Wei-Chieh and Li, Yangning and Fang, Liancheng and Wang, Zhen and Yu, Philip S},
  journal={Findings of the Association for Computational Linguistics: EMNLP},
  year={2025}
}

@article{zhang2025openread,
  title={OpenREAD: Reinforced Open-Ended Reasoing for End-to-End Autonomous Driving with LLM-as-Critic},
  author={Zhang, Songyan and Huang, Wenhui and Chen, Zhan and Collister, Chua Jiahao and Huang, Qihang and Lv, Chen},
  journal={arXiv preprint arXiv:2512.01830},
  year={2025}
}

@article{cui2023drivellm,
  title={Drivellm: Charting the path toward full autonomous driving with large language models},
  author={Cui, Yaodong and Huang, Shucheng and Zhong, Jiaming and Liu, Zhenan and Wang, Yutong and Sun, Chen and Li, Bai and Wang, Xiao and Khajepour, Amir},
  journal={IEEE Transactions on Intelligent Vehicles},
  year={2023},
}

@inproceedings{yang2025trajectory,
  title={Trajectory-llm: A language-based data generator for trajectory prediction in autonomous driving},
  author={Yang, Kairui and Guo, Zihao and Lin, Gengjie and Dong, Haotian and Huang, Zhao and Wu, Yipeng and Zuo, Die and Peng, Jibin and Zhong, Ziyuan and Wang, Xin and others},
  booktitle={The Thirteenth International Conference on Learning Representations},
  year={2025}
}

@article{xu2025tell,
  title={Tell-drive: Enhancing autonomous driving with teacher llm-guided deep reinforcement learning},
  author={Xu, Chengkai and Liu, Jiaqi and Fang, Shiyu and Cui, Yiming and Chen, Dong and Hang, Peng and Sun, Jian},
  journal={arXiv preprint arXiv:2502.01387},
  year={2025}
}

@article{shao2024deepseekmath,
  title={Deepseekmath: Pushing the limits of mathematical reasoning in open language models},
  author={Shao, Zhihong and Wang, Peiyi and Zhu, Qihao and Xu, Runxin and Song, Junxiao and Bi, Xiao and Zhang, Haowei and Zhang, Mingchuan and Li, YK and Wu, Yang and others},
  journal={arXiv preprint arXiv:2402.03300},
  year={2024}
}

@article{rafailov2023direct,
  title={Direct preference optimization: Your language model is secretly a reward model},
  author={Rafailov, Rafael and Sharma, Archit and Mitchell, Eric and Manning, Christopher D and Ermon, Stefano and Finn, Chelsea},
  journal={Advances in neural information processing systems},
  year={2023}
}

@article{bai2025qwen2,
  title={Qwen2. 5-vl technical report},
  author={Bai, Shuai and Chen, Keqin and Liu, Xuejing and Wang, Jialin and Ge, Wenbin and Song, Sibo and Dang, Kai and Wang, Peng and Wang, Shijie and Tang, Jun and others},
  journal={arXiv preprint arXiv:2502.13923},
  year={2025}
}

@misc{openscene2023,
      title = {OpenScene: The Largest Up-to-Date 3D Occupancy Prediction Benchmark in Autonomous Driving},
      author = {OpenScene Contributors},
      howpublished={\url{https://github.com/OpenDriveLab/OpenScene}},
      year = {2023}
}

@article{sima2023_occnet,
      title={Scene as Occupancy}, 
      author={Chonghao Sima and Wenwen Tong and Tai Wang and Li Chen and Silei Wu and Hanming Deng  and Yi Gu and Lewei Lu and Ping Luo and Dahua Lin and Hongyang Li},
      year={2023},
}

@inproceedings{ma2024learning,
  title={Learning autonomous driving tasks via human feedbacks with large language models},
  author={Ma, Yunsheng and Cao, Xu and Ye, Wenqian and Cui, Can and Mei, Kai and Wang, Ziran},
  booktitle={Findings of the Association for Computational Linguistics: EMNLP 2024},
  year={2024}
}

@article{tian2024drivevlm,
  title={Drivevlm: The convergence of autonomous driving and large vision-language models},
  author={Tian, Xiaoyu and Gu, Junru and Li, Bailin and Liu, Yicheng and Wang, Yang and Zhao, Zhiyong and Zhan, Kun and Jia, Peng and Lang, Xianpeng and Zhao, Hang},
  journal={arXiv preprint arXiv:2402.12289},
  year={2024}
}

@article{li2025drive,
  title={Drive-R1: Bridging Reasoning and Planning in VLMs for Autonomous Driving with Reinforcement Learning},
  author={Li, Yue and Tian, Meng and Zhu, Dechang and Zhu, Jiangtong and Lin, Zhenyu and Xiong, Zhiwei and Zhao, Xinhai},
  journal={arXiv preprint arXiv:2506.18234},
  year={2025}
}

@inproceedings{shao2024lmdrive,
  title={Lmdrive: Closed-loop end-to-end driving with large language models},
  author={Shao, Hao and Hu, Yuxuan and Wang, Letian and Song, Guanglu and Waslander, Steven L and Liu, Yu and Li, Hongsheng},
  booktitle={Proceedings of the IEEE/CVF Conference on Computer Vision and Pattern Recognition},
  year={2024}
}

@article{zhu2025internvl3,
  title={Internvl3: Exploring advanced training and test-time recipes for open-source multimodal models},
  author={Zhu, Jinguo and Wang, Weiyun and Chen, Zhe and Liu, Zhaoyang and Ye, Shenglong and Gu, Lixin and Tian, Hao and Duan, Yuchen and Su, Weijie and Shao, Jie and others},
  journal={arXiv preprint arXiv:2504.10479},
  year={2025}
}

@article{zhou2025autovla,
  title={AutoVLA: A Vision-Language-Action Model for End-to-End Autonomous Driving with Adaptive Reasoning and Reinforcement Fine-Tuning},
  author={Zhou, Zewei and Cai, Tianhui and Zhao, Seth Z and Zhang, Yun and Huang, Zhiyu and Zhou, Bolei and Ma, Jiaqi},
  journal={arXiv preprint arXiv:2506.13757},
  year={2025}
}

@article{zhou2025opendrivevla,
  title={Opendrivevla: Towards end-to-end autonomous driving with large vision language action model},
  author={Zhou, Xingcheng and Han, Xuyuan and Yang, Feng and Ma, Yunpu and Knoll, Alois C},
  journal={arXiv preprint arXiv:2503.23463},
  year={2025}
}

@article{li2025recogdrive,
  title={Recogdrive: A reinforced cognitive framework for end-to-end autonomous driving},
  author={Li, Yongkang and Xiong, Kaixin and Guo, Xiangyu and Li, Fang and Yan, Sixu and Xu, Gangwei and Zhou, Lijun and Chen, Long and Sun, Haiyang and Wang, Bing and others},
  journal={arXiv preprint arXiv:2506.08052},
  year={2025}
}

@article{jiang2025alphadrive,
  title={Alphadrive: Unleashing the power of vlms in autonomous driving via reinforcement learning and reasoning},
  author={Jiang, Bo and Chen, Shaoyu and Zhang, Qian and Liu, Wenyu and Wang, Xinggang},
  journal={arXiv preprint arXiv:2503.07608},
  year={2025}
}

@article{driess2023palm,
  title={Palm-e: An embodied multimodal language model},
  author={Driess, Danny and Xia, Fei and Sajjadi, Mehdi SM and Lynch, Corey and Chowdhery, Aakanksha and Wahid, Ayzaan and Tompson, Jonathan and Vuong, Quan and Yu, Tianhe and Huang, Wenlong and others},
  journal={arXiv preprint arXiv:2303.03378},
  year={2023}
}

@article{luo2025adathinkdrive,
  title={AdaThinkDrive: Adaptive Thinking via Reinforcement Learning for Autonomous Driving},
  author={Luo, Yuechen and Li, Fang and Xu, Shaoqing and Lai, Zhiyi and Yang, Lei and Chen, Qimao and Luo, Ziang and Xie, Zixun and Jiang, Shengyin and Liu, Jiaxin and others},
  journal={arXiv preprint arXiv:2509.13769},
  year={2025}
}

@article{zheng2025driveagent,
  title={DriveAgent-R1: Advancing VLM-based Autonomous Driving with Active Perception and Hybrid Thinking},
  author={Zheng, Weicheng and Mao, Xiaofei and Ye, Nanfei and Li, Pengxiang and Zhan, Kun and Lang, Xianpeng and Zhao, Hang},
  journal={arXiv preprint arXiv:2507.20879},
  year={2025}
}

@article{fang2025corevla,
  title={CoReVLA: A Dual-Stage End-to-End Autonomous Driving Framework for Long-Tail Scenarios via Collect-and-Refine},
  author={Fang, Shiyu and Cui, Yiming and Liang, Haoyang and Lv, Chen and Hang, Peng and Sun, Jian},
  journal={arXiv preprint arXiv:2509.15968},
  year={2025}
}

@article{wang2025cogad,
  title={CogAD: Cognitive-Hierarchy Guided End-to-End Autonomous Driving},
  author={Wang, Zhennan and Teng, Jianing and Xiang, Canqun and Chen, Kangliang and Pan, Xing and Deng, Lu and Gu, Weihao},
  journal={arXiv preprint arXiv:2505.21581},
  year={2025}
}

@inproceedings{lu2025real,
  title={ReAL-AD: Towards Human-Like Reasoning in End-to-End Autonomous Driving},
  author={Lu, Yuhang and Tu, Jiadong and Ma, Yuexin and Zhu, Xinge},
  booktitle={Proceedings of the IEEE/CVF International Conference on Computer Vision},
  year={2025}
}

@article{mao2023language,
  title={A language agent for autonomous driving},
  author={Mao, Jiageng and Ye, Junjie and Qian, Yuxi and Pavone, Marco and Wang, Yue},
  journal={arXiv preprint arXiv:2311.10813},
  year={2023}
}

@article{hwang2024emma,
  title={Emma: End-to-end multimodal model for autonomous driving},
  author={Hwang, Jyh-Jing and Xu, Runsheng and Lin, Hubert and Hung, Wei-Chih and Ji, Jingwei and Choi, Kristy and Huang, Di and He, Tong and Covington, Paul and Sapp, Benjamin and others},
  journal={arXiv preprint arXiv:2410.23262},
  year={2024}
}

@inproceedings{pan2024vlp,
  title={Vlp: Vision language planning for autonomous driving},
  author={Pan, Chenbin and Yaman, Burhaneddin and Nesti, Tommaso and Mallik, Abhirup and Allievi, Alessandro G and Velipasalar, Senem and Ren, Liu},
  booktitle={Proceedings of the IEEE/CVF Conference on Computer Vision and Pattern Recognition},
  year={2024}
}

@article{yang2023llm4drive,
  title={Llm4drive: A survey of large language models for autonomous driving},
  author={Yang, Zhenjie and Jia, Xiaosong and Li, Hongyang and Yan, Junchi},
  journal={arXiv preprint arXiv:2311.01043},
  year={2023}
}

@inproceedings{zhang2025adadrive,
  title={AdaDrive: Self-Adaptive Slow-Fast System for Language-Grounded Autonomous Driving},
  author={Zhang, Ruifei and Xie, Junlin and Zhang, Wei and Chen, Weikai and Tan, Xiao and Wan, Xiang and Li, Guanbin},
  booktitle={Proceedings of the IEEE/CVF International Conference on Computer Vision},
  year={2025}
}

@inproceedings{zhang2025vldrive,
  title={VLDrive: Vision-Augmented Lightweight MLLMs for Efficient Language-grounded Autonomous Driving},
  author={Zhang, Ruifei and Zhang, Wei and Tan, Xiao and Yang, Sibei and Wan, Xiang and Luo, Xiaonan and Li, Guanbin},
  booktitle={Proceedings of the IEEE/CVF International Conference on Computer Vision},
  year={2025}
}

@article{zeng2025futuresightdrive,
  title={Futuresightdrive: Thinking visually with spatio-temporal cot for autonomous driving},
  author={Zeng, Shuang and Chang, Xinyuan and Xie, Mengwei and Liu, Xinran and Bai, Yifan and Pan, Zheng and Xu, Mu and Wei, Xing and Guo, Ning},
  journal={arXiv preprint arXiv:2505.17685},
  year={2025}
}

@article{zhang2025safeauto,
  title={Safeauto: Knowledge-enhanced safe autonomous driving with multimodal foundation models},
  author={Zhang, Jiawei and Yang, Xuan and Wang, Taiqi and Yao, Yu and Petiushko, Aleksandr and Li, Bo},
  journal={arXiv preprint arXiv:2503.00211},
  year={2025}
}

@inproceedings{xie2025s4,
  title={S4-Driver: Scalable Self-Supervised Driving Multimodal Large Language Model with Spatio-Temporal Visual Representation},
  author={Xie, Yichen and Xu, Runsheng and He, Tong and Hwang, Jyh-Jing and Luo, Katie and Ji, Jingwei and Lin, Hubert and Chen, Letian and Lu, Yiren and Leng, Zhaoqi and others},
  booktitle={Proceedings of the Computer Vision and Pattern Recognition Conference},
  year={2025}
}

@article{yang2025drivemoe,
  title={DriveMoE: Mixture-of-Experts for Vision-Language-Action Model in End-to-End Autonomous Driving},
  author={Yang, Zhenjie and Chai, Yilin and Jia, Xiaosong and Li, Qifeng and Shao, Yuqian and Zhu, Xuekai and Su, Haisheng and Yan, Junchi},
  journal={arXiv preprint arXiv:2505.16278},
  year={2025}
}

@article{yuan2024rag,
  title={Rag-driver: Generalisable driving explanations with retrieval-augmented in-context learning in multi-modal large language model},
  author={Yuan, Jianhao and Sun, Shuyang and Omeiza, Daniel and Zhao, Bo and Newman, Paul and Kunze, Lars and Gadd, Matthew},
  journal={arXiv preprint arXiv:2402.10828},
  year={2024}
}

@inproceedings{sima2024drivelm,
  title={Drivelm: Driving with graph visual question answering},
  author={Sima, Chonghao and Renz, Katrin and Chitta, Kashyap and Chen, Li and Zhang, Hanxue and Xie, Chengen and Bei{\ss}wenger, Jens and Luo, Ping and Geiger, Andreas and Li, Hongyang},
  booktitle={European conference on computer vision},
  year={2024},
}

@article{liu2025hybrid,
  title={Hybrid-prediction integrated planning for autonomous driving},
  author={Liu, Haochen and Huang, Zhiyu and Huang, Wenhui and Yang, Haohan and Mo, Xiaoyu and Lv, Chen},
  journal={IEEE Transactions on Pattern Analysis and Machine Intelligence},
  year={2025},
}

@inproceedings{han2025dme,
  title={Dme-driver: Integrating human decision logic and 3d scene perception in autonomous driving},
  author={Han, Wencheng and Guo, Dongqian and Xu, Cheng-Zhong and Shen, Jianbing},
  booktitle={Proceedings of the AAAI Conference on Artificial Intelligence},
  year={2025}
}

@article{mao2023gpt,
  title={Gpt-driver: Learning to drive with gpt},
  author={Mao, Jiageng and Qian, Yuxi and Ye, Junjie and Zhao, Hang and Wang, Yue},
  journal={arXiv preprint arXiv:2310.01415},
  year={2023}
}

@article{xu2024drivegpt4,
  title={Drivegpt4: Interpretable end-to-end autonomous driving via large language model},
  author={Xu, Zhenhua and Zhang, Yujia and Xie, Enze and Zhao, Zhen and Guo, Yong and Wong, Kwan-Yee K and Li, Zhenguo and Zhao, Hengshuang},
  journal={IEEE Robotics and Automation Letters},
  year={2024},
}

@inproceedings{Dauner2024NEURIPS,
	title = {NAVSIM: Data-Driven Non-Reactive Autonomous Vehicle Simulation and Benchmarking},
	author = {Daniel Dauner and Marcel Hallgarten and Tianyu Li and Xinshuo Weng and Zhiyu Huang and Zetong Yang and Hongyang Li and Igor Gilitschenski and Boris Ivanovic and Marco Pavone and Andreas Geiger and Kashyap Chitta},
	booktitle = {Advances in Neural Information Processing Systems (NeurIPS)},
	year = {2024},
}

@article{yuan2025autodrive,
  title={AutoDrive-R$^{2}$: Incentivizing Reasoning and Self-Reflection Capacity for VLA Model in Autonomous Driving},
  author={Yuan, Zhenlong and Tang, Jing and Luo, Jinguo and Chen, Rui and Qian, Chengxuan and Sun, Lei and Chu, Xiangxiang and Cai, Yujun and Zhang, Dapeng and Li, Shuo},
  journal={arXiv preprint arXiv:2509.01944},
  year={2025}
}

@article{yan2025rlgf,
  title={RLGF: Reinforcement Learning with Geometric Feedback for Autonomous Driving Video Generation},
  author={Yan, Tianyi and Han, Wencheng and Zhou, Xia and Zhang, Xueyang and Zhan, Kun and Xu, Cheng-zhong and Shen, Jianbing},
  journal={arXiv preprint arXiv:2509.16500},
  year={2025}
}

@article{ma2024survey,
  title={A survey on vision-language-action models for embodied ai},
  author={Ma, Yueen and Song, Zixing and Zhuang, Yuzheng and Hao, Jianye and King, Irwin},
  journal={arXiv preprint arXiv:2405.14093},
  year={2024}
}

@article{fu2025orion,
  title={Orion: A holistic end-to-end autonomous driving framework by vision-language instructed action generation},
  author={Fu, Haoyu and Zhang, Diankun and Zhao, Zongchuang and Cui, Jianfeng and Liang, Dingkang and Zhang, Chong and Zhang, Dingyuan and Xie, Hongwei and Wang, Bing and Bai, Xiang},
  journal={arXiv preprint arXiv:2503.19755},
  year={2025}
}

@article{cao2025pseudo,
  title={Pseudo-simulation for autonomous driving},
  author={Cao, Wei and Hallgarten, Marcel and Li, Tianyu and Dauner, Daniel and Gu, Xunjiang and Wang, Caojun and Miron, Yakov and Aiello, Marco and Li, Hongyang and Gilitschenski, Igor and others},
  journal={arXiv preprint arXiv:2506.04218},
  year={2025}
}

@inproceedings{mirzaie2025interpretable,
  title={Interpretable Decision-Making for End-to-End Autonomous Driving},
  author={Mirzaie, Mona and Rosenhahn, Bodo},
  booktitle={Proceedings of the IEEE/CVF International Conference on Computer Vision},
  year={2025}
}

@inproceedings{sun2025sparsedrive,
  title={Sparsedrive: End-to-end autonomous driving via sparse scene representation},
  author={Sun, Wenchao and Lin, Xuewu and Shi, Yining and Zhang, Chuang and Wu, Haoran and Zheng, Sifa},
  booktitle={2025 IEEE International Conference on Robotics and Automation (ICRA)},
  year={2025},
}

@article{gao2024vista,
  title={Vista: A generalizable driving world model with high fidelity and versatile controllability},
  author={Gao, Shenyuan and Yang, Jiazhi and Chen, Li and Chitta, Kashyap and Qiu, Yihang and Geiger, Andreas and Zhang, Jun and Li, Hongyang},
  journal={Advances in Neural Information Processing Systems},
  volume={37},
  pages={91560--91596},
  year={2024}
}

@article{chen2024end,
  title={End-to-end autonomous driving: Challenges and frontiers},
  author={Chen, Li and Wu, Penghao and Chitta, Kashyap and Jaeger, Bernhard and Geiger, Andreas and Li, Hongyang},
  journal={IEEE Transactions on Pattern Analysis and Machine Intelligence},
  year={2024},
}

@inproceedings{chen2025drivinggpt,
  title={Drivinggpt: Unifying driving world modeling and planning with multi-modal autoregressive transformers},
  author={Chen, Yuntao and Wang, Yuqi and Zhang, Zhaoxiang},
  booktitle={Proceedings of the IEEE/CVF International Conference on Computer Vision},
  year={2025}
}

@inproceedings{jiang2023vad,
  title={Vad: Vectorized scene representation for efficient autonomous driving},
  author={Jiang, Bo and Chen, Shaoyu and Xu, Qing and Liao, Bencheng and Chen, Jiajie and Zhou, Helong and Zhang, Qian and Liu, Wenyu and Huang, Chang and Wang, Xinggang},
  booktitle={Proceedings of the IEEE/CVF International Conference on Computer Vision},
  year={2023}
}

@article{nuscenes2019,
  title={nuScenes: A multimodal dataset for autonomous driving},
  author={Holger Caesar and Varun Bankiti and Alex H. Lang and Sourabh Vora and 
          Venice Erin Liong and Qiang Xu and Anush Krishnan and Yu Pan and 
          Giancarlo Baldan and Oscar Beijbom},
  journal={arXiv preprint arXiv:1903.11027},
  year={2019}
}

@inproceedings{zheng2024genad,
  title={Genad: Generative end-to-end autonomous driving},
  author={Zheng, Wenzhao and Song, Ruiqi and Guo, Xianda and Zhang, Chenming and Chen, Long},
  booktitle={European Conference on Computer Vision},
  year={2024},
}

@inproceedings{liao2025diffusiondrive,
  title={Diffusiondrive: Truncated diffusion model for end-to-end autonomous driving},
  author={Liao, Bencheng and Chen, Shaoyu and Yin, Haoran and Jiang, Bo and Wang, Cheng and Yan, Sixu and Zhang, Xinbang and Li, Xiangyu and Zhang, Ying and Zhang, Qian and others},
  booktitle={Proceedings of the Computer Vision and Pattern Recognition Conference},
  year={2025}
}

@article{jia2024bench2drive,
  title={Bench2drive: Towards multi-ability benchmarking of closed-loop end-to-end autonomous driving},
  author={Jia, Xiaosong and Yang, Zhenjie and Li, Qifeng and Zhang, Zhiyuan and Yan, Junchi},
  journal={Advances in Neural Information Processing Systems},
  year={2024}
}

@inproceedings{dosovitskiy2017carla,
  title={CARLA: An open urban driving simulator},
  author={Dosovitskiy, Alexey and Ros, German and Codevilla, Felipe and Lopez, Antonio and Koltun, Vladlen},
  booktitle={Conference on robot learning},
  year={2017},
}
\clearpage
\appendix

\section{Related Work}
\label{app:related_work}

\subsection{End-to-end Autonomous Driving}
End-to-end (E2E) autonomous driving has emerged as a dominant paradigm, aiming to replace the traditional modular pipeline of perception, prediction, and planning with a single, jointly optimized neural network~\cite{chen2024vadv2, chen2024end, zhang2025safeauto, xie2025s4, zhang2025vldrive, mirzaie2025interpretable}. 
This approach directly maps raw sensor inputs, such as images and LiDAR point clouds, to driving commands or future trajectories, thereby minimizing error accumulation across cascaded modules.
Pioneering works in this domain have demonstrated the efficacy of integrating multiple intermediate tasks into a unified, planning-oriented framework~\cite{chen2024end, sun2025sparsedrive, yan2025rlgf}. 
For instance, UniAD~\cite{hu2023planning} and VAD~\cite{jiang2023vad} established strong baselines by holistically modeling the driving scene and formulating the entire system as a planning-centric problem. 
These methods showcase impressive performance by leveraging shared representations across perception and prediction to benefit the final planning output.
However, deterministic trajectory regression fails to capture the inherent uncertainty and multi-modal nature of real-world driving scenarios. 
To address this, recent efforts have shifted towards modeling a distribution over possible future actions. 
VADv2~\cite{chen2024vadv2} introduced probabilistic planning by predicting a probability distribution over a set of actions and sampling from it to control the vehicle. 
Concurrently, a new paradigm employing generative models has gained traction. 
Methods like GenAD~\cite{zheng2024genad} and DiffusionDrive~\cite{liao2025diffusiondrive} leverage the power of diffusion models to generate diverse and plausible multi-modal trajectories, better reflecting the complex decision-making process of human drivers.

Despite promising results on open-loop benchmarks, where models predict trajectories on prerecorded logs, a significant performance gap often exists when these systems are deployed in interactive, closed-loop simulations. 
As highlighted by recent studies~\cite{jia2024bench2drive, cao2025pseudo}, many E2E models tend to overfit to the ego-vehicle's status in the training data, leading to suboptimal or unsafe behaviors in dynamic, reactive environments~\cite{fu2025orion}. 
While some works adopt closed-loop evaluation in simulators like CARLA~\cite{dosovitskiy2017carla}, their performance often reveals the brittleness of models trained primarily for open-loop metrics.
This discrepancy underscores the critical need for developing E2E frameworks that are not only accurate in offline evaluation but also robust and reliable in realistic, interactive driving scenarios. 
To address these limitations, several recent approaches have successfully employed Vision-Language-Action (VLA) models, demonstrating strong closed-loop performance in autonomous driving systems~\cite{fu2025orion, li2025recogdrive, zhou2025opendrivevla, chen2025drivinggpt}.

\begin{figure*}[t]
    \centering
    \includegraphics[width=\linewidth]{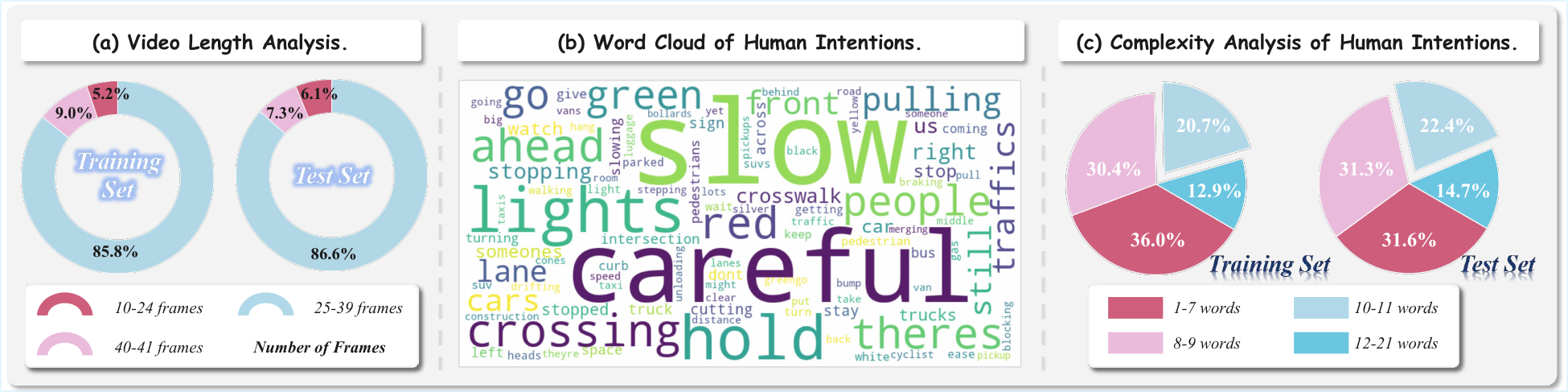}
    \vspace{-7mm}
     \caption{\small \textbf{Statistical analysis of the Intention-Drive benchmark.} 
    (a) Distribution of video frame counts, ensuring coverage of diverse temporal horizons.
    (b) Word cloud of human intentions illustrating semantic richness.
    (c) Distribution of instruction lengths, highlighting the linguistic complexity with a significant portion of long, detailed intentions.}
    \label{fig:dataset_analysis}
    \vspace{-4mm}
\end{figure*}

\subsection{Vision-Language-Action Model}
Vision-Language-Action (VLA) models represent a paradigm shift in creating intelligent agents, aiming to bridge the gap between multi-modal perception, natural language understanding, and physical action generation in the real world~\cite{ma2024survey, zhou2025autovla, li2025drive}. 
Originating from the field of robotics, VLAs are designed to interpret high-level instructions and execute them by generating a sequence of low-level actions, making them a natural architectural choice for intention-driven autonomous driving~\cite{driess2023palm, tian2024drivevlm, mao2023gpt, zhang2025adadrive, zhang2025openread, cui2023drivellm}. 
These models move beyond passive scene description and towards active interaction, which is critical for fulfilling complex human intentions in dynamic environments like driving.

In the context of autonomous driving, a significant body of work aims to emulate human-like cognitive structures to make decision-making more robust and interpretable~\cite{xu2025tell}. 
For instance, models like CogAD~\cite{wang2025cogad} and ReAL-AD~\cite{lu2025real} explicitly design hierarchical frameworks inspired by cognitive psychology, breaking down the planning process from high-level intention or strategy to fine-grained trajectory execution. 
This structured approach is often enhanced by sophisticated reasoning mechanisms. 
AutoDrive-R2~\cite{yuan2025autodrive} incorporates a chain-of-thought process with self-reflection, while ReCogDrive~\cite{li2025recogdrive} integrates a cognitive VLM with a diffusion planner to address the modality mismatch between discrete language commands and continuous driving actions.
To move beyond the limitations of simple imitation learning from static datasets, reinforcement learning (RL) has emerged as a key technique for policy refinement. 
AlphaDrive~\cite{jiang2025alphadrive} leverages Group Relative Policy Optimization (GRPO)~\cite{shao2024deepseekmath} with planning-oriented rewards to significantly boost performance. 
Similarly, AdaThinkDrive~\cite{luo2025adathinkdrive} employs RL to train an agent that can adaptively switch between ``fast'' direct prediction and ``slow'' deliberative reasoning based on scene complexity. 
Other approaches focus on learning from more nuanced human feedback; CoReVLA~\cite{fang2025corevla}, for example, uses Direct Preference Optimization (DPO)~\cite{rafailov2023direct} to refine its model from sparse human takeover data in long-tail scenarios. 
Pushing the frontier even further, DriveAgent-R1~\cite{zheng2025driveagent} pioneers the concept of active perception, where the agent can proactively use vision tools to seek out additional information to resolve uncertainty, representing a paradigm shift from passive observation to active, grounded decision-making.

\section{Dataset Analysis}

To better understand the challenges posed by Intention-Drive, we provide a comprehensive statistical analysis of the dataset characteristics, as illustrated in Figure~\ref{fig:dataset_analysis}. 
The analysis focuses on three key dimensions: temporal diversity, semantic richness, and linguistic complexity.

\noindent\textbf{Temporal Diversity.}
Figure~\ref{fig:dataset_analysis}(a) presents the distribution of video sequence lengths across the training and test sets. 
The dataset covers a broad range of temporal horizons, reflecting the variable duration of real-world driving maneuvers. 
The distribution indicates that the dataset is not limited to short-term reactions but includes extended sequences that require the agent to maintain temporal consistency over longer periods. 

\noindent\textbf{Semantic Richness.}
Unlike traditional datasets restricted to steering commands, our dataset emphasizes the behavioral and interactive aspects of driving. 
The word cloud in Figure~\ref{fig:dataset_analysis}(b) highlights high-frequency keywords such as ``slow'', ``careful'', ``traffic'', ``wait'', and ``pedestrian''. 
The prominence of these terms underscores the shift from command-following to intention-fulfillment. 

\noindent\textbf{Linguistic Complexity.}
Figure~\ref{fig:dataset_analysis}(c) analyzes the complexity of human intentions based on instruction length. 
Unlike conventional datasets dominated by short phrases, our benchmark features a substantial distribution of long-form instructions. 
Specifically, instructions exceeding 7 words constitute the majority. 
This linguistic richness necessitates strong language grounding capabilities, as the E2E AD model must parse the sentence to extract the destination, behavioral modifiers, and safety constraints embedded in the driver's intention.

\begin{figure*}[t]
    \centering
    \includegraphics[width=\linewidth]{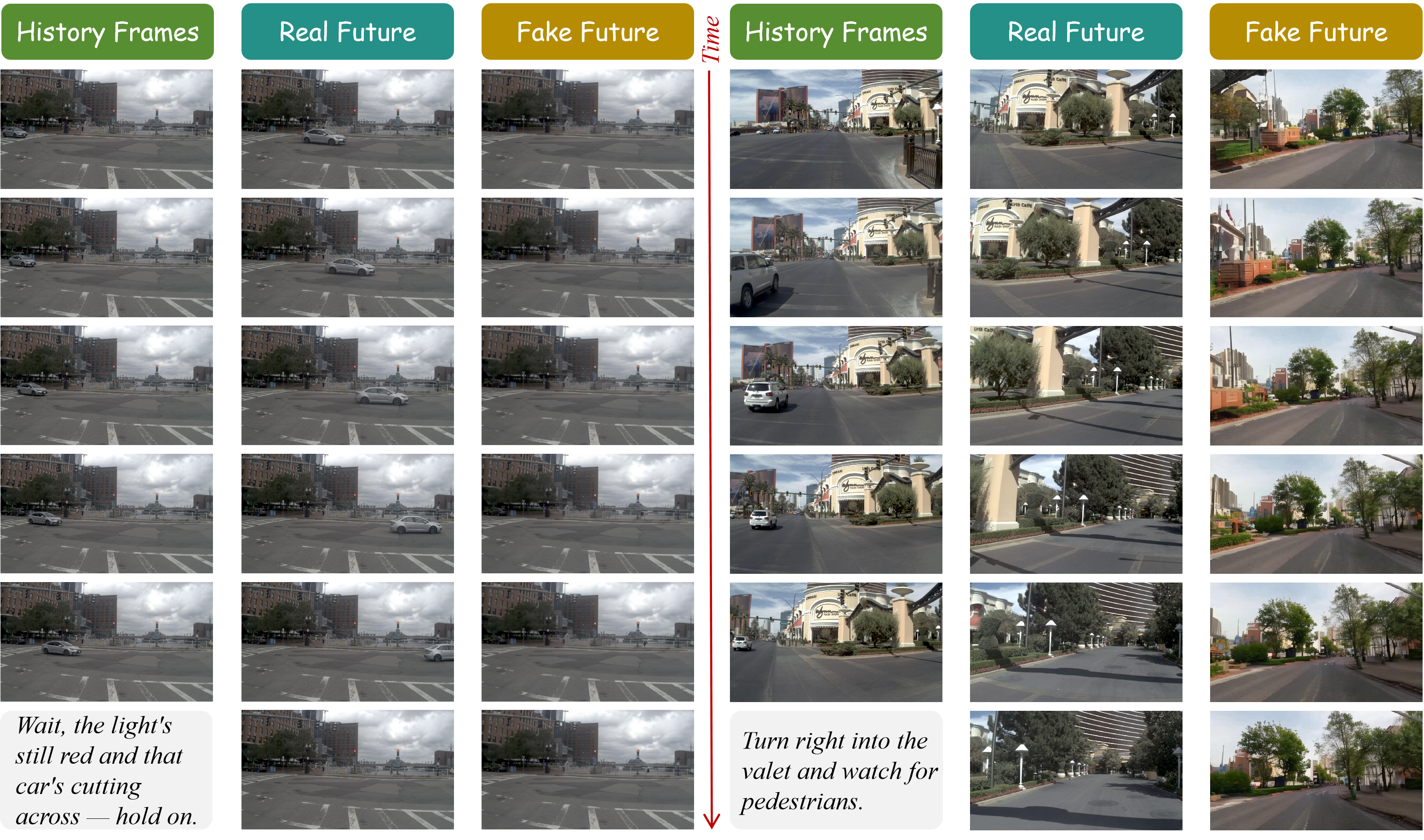}
    \vspace{-7mm}
     \caption{\small \textbf{Case study for InternVL3.0-2B*.} We display the historical context, the ground truth future (Real Future), and the hallucinated future (Fake Future) generated by the world model conditioned on our agent's predicted trajectory. The comparison highlights that the generated "Fake Future" successfully reflects the semantic constraints of the human instructions—such as holding for a cutting vehicle (left) or identifying a specific valet entrance (right)—thereby verifying the semantic consistency of the planned action beyond simple geometric metrics.}
    \label{fig:case}
    \vspace{-4mm}
\end{figure*}

\section{Case Study}

To qualitatively validate the effectiveness of our framework and the proposed Imagined Future Alignment (IFA) protocol, we present a visualization of the generative evaluation process in Figure \ref{fig:case}. This figure demonstrates how the pre-trained generative world model "hallucinates" the future consequences of the agent's predicted trajectory, enabling a direct semantic comparison against the ground truth (Real Future) and the user's high-level instruction. We select two scenarios to illustrate the model's capability in handling safety-critical constraints and specific navigational goals.

In the left panel, the user issues a complex, safety-oriented command: \textit{Wait, the light's still red and that car's cutting across — hold on.} This instruction requires the agent to prioritize immediate safety over progress and recognize the dynamic hazard. The visualized "Fake Future" confirms that the agent successfully generated a stationary trajectory. Crucially, because the trajectory was correct, the world model renders the crossing vehicle passing safely in front of the ego-vehicle, mirroring the dynamics observed in the "Real Future." The Semantic Judge (VLM) evaluating this hallucinated clip can thus verify that the "hold on" condition was met, resulting in a high Score.

Conversely, the right panel depicts a goal-oriented scenario where the driver instructs: \textit{Turn right into the valet and watch for pedestrians.} The challenge here lies in grounding the semantic entity "valet" to the correct spatial location rather than a generic road intersection. The "Fake Future" visualization shows the perspective shifting smoothly into the specific driveway, closely aligning with the visual cues in the ground truth video. This demonstrates that the agent's planned trajectory effectively guided the world model to render the correct destination. These visualizations highlight that our Intention-Driven framework does not merely regress coordinates but semantically fulfills the human's underlying intent, which is faithfully captured and verified by the IFA protocol.

\end{document}